# A Fast Compression-based Similarity Measure with Applications to Content-based Image Retrieval

*Daniele Cerra\* and Mihai Datcu*

German Aerospace Center (DLR), Earth Observation Center (EOC), Münchnerstr. 20, 82234 Wessling, Germany

\*Corresponding author: Phone + 49 8153 28 1496; Fax + 49 8153 28 1444; E-Mail daniele.cerra@dlr.de

*Abstract*— Compression-based similarity measures are effectively employed in applications on diverse data types with a basically parameter-free approach. Nevertheless, there are problems in applying these techniques to medium-to-large datasets which have been seldom addressed. This paper proposes a similarity measure based on compression with dictionaries, the Fast Compression Distance (FCD), which reduces the complexity of these methods, without degradations in performance. On its basis a content-based color image retrieval system is defined, which can be compared to state-of-the-art methods based on invariant color features. Through the FCD a better understanding of compression-based techniques is achieved, by performing experiments on datasets which are larger than the ones analyzed so far in literature.

*Index Terms*— Compression, similarity measure, information retrieval.

## 1. INTRODUCTION

Compression-based similarity measures employ in an unusual way general off-the-shelf compressors, by exploiting them to estimate the amount of information shared by any two objects. Such techniques, of which the most well-known is the Normalized Compression Distance (NCD) [1], have a characteristic parameter-free flavor, which decreases the disadvantages of working with parameter-dependent algorithms [2] [3]. In addition, the data-driven approach characteristic of these notions permits to apply them to any kind of data, and in several domains such as unsupervised clustering, classification and anomaly detection [4].

Nevertheless these data-driven techniques have the drawback of being computationally intensive and have been applied, in the general case, to restricted datasets: in [4] the authors estimate the running time of a variant of NCD as "less than ten seconds (on a 2.65 GHz machine) to process a million data points". In the case of images datasets, almost ten seconds would be then needed to process five RGB images of size 256x256. This represents a major drawback for compression-based analysis in real applications, where usually medium-to-large datasets are involved. In [7] the authors suggest to perform a first dimensionality reduction step using standard clustering procedures, computing then with NCD a distance matrix related to the clusters instead of all the objects in the dataset. In [13] a different solution is brought forward by suggesting a Support Vector Machine (SVM) [14] based classification, where NCD distances with representative objects for each class, which are chosen as anchors, form a feature vector which is then used as an input for the SVM. Nevertheless, this solution introduces undesired subjective choices, such as choosing the right anchors. Results would then be based on a partial analysis of the dataset, and this would be a drawback especially for retrieval tasks, where a decision has to be taken for each object in the set.

Finally, the Pattern Representation using Data Compression (PRDC) [5], a general classification methodology based on the compression with dictionaries directly extracted from the data, is faster but less effective than NCD [17].

This paper defines a compression-based similarity measure, the Fast Compression Distance (FCD), which combines the accuracy of NCD with the reduced complexity of PRDC. In a first offline step, the images are quantized in a convenient color space and converted into strings, after being modified to preserve some textural information in the process; subsequently, representative dictionaries are extracted from each object and similarities between individual images are computed by comparing each couple of dictionaries. This allows testing for the first time compression-based techniques on datasets composed of up to 10,000 images.

A Content-based Image Retrieval (CBIR) system based on FCD is then defined, and retrieval results suggest that FCD is comparable to state of the art methods based on invariant color features, and outperforms other compression-based methods.

The next sections of this paper are structured as follows. We give a reminder on compression-based similarity measures in Section 2. Section 3 introduces the FCD and on its basis defines a CBIR system. Experiments on 4 datasets and comparisons to other methods are presented in Section 4. We conclude in Section 5.

## 2. COMPRESSION-BASED SIMILARITY MEASURES

The most widely known and used compression based similarity measure for general data is the Normalized Compression Distance (NCD), proposed by Li et al. [1]. The NCD derives from the Kolmogorov complexity $K(x)$ of an object $x$, which quantifies how difficult it is to compute or describe $x$ [6]: the quantity $K(x)$ is incomputable, but can be approximated by compression algorithms and on its basis the NCD is defined for any two objects $x$ and $y$ as:

$$NCD(x,y) = \frac{C(x,y) - \min\{C(x),C(y)\}}{\max\{C(x),C(y)\}} \quad (1)$$

where $C(x)$ represents the size of the (lossless) compressed version of $x$, and $C(x,y)$ the size of the file obtained by compressing the concatenation of $x$ and $y$. The idea is that if $x$ and $y$ share common information they will compress better together than separately, as the compressor will be able to reuse the recurring patterns found in one of them to more efficiently compress the other. The *NCD* can be explicitly computed between any two strings or files $x$ and $y$ and has a characteristic data-driven, parameter-free approach, that allows performing clustering, classification and anomaly detection on diverse data types [7-12].

The problems in applying NCD-like distances to large datasets have been seldom addressed. Usually, the data-driven approach of these methods requires iterative processing of the full data, not allowing compact representations in any explicit parameter space. Most of the experiments performed on the basis of these techniques include the computation of a distance matrix between all objects in a dataset, with the latter seldom containing more than 100 objects (see e.g. [4] [7]).

Another compression-based technique seems more apt to be exploited for these means: the Pattern Representation based on Data Compression (*PRDC*), a classification methodology introduced by Watanabe et al. [5] independently from the *NCD*. A link between these two quantities is investigated in [15]. The idea to the basis of *PRDC* is to extract offline typical dictionaries, obtained with a compressor belonging to the LZ family [16], directly from the data previously encoded into strings: these dictionaries are later used to compress other files in order to discover similarities with them on the basis of the dictionaries compression power. For two strings *x* and *y* PRDC is usually faster than *NCD*, as the joint compression of *x* and *y* which is the most computationally intensive step is avoided. Furthermore, if *y* is compared to multiple objects, the compression of *y*, implicitly carried out by extracting the dictionary *D*(*y*), has to be computed only once, while NCD always processes from scratch the full *x* and *y* in the computation of each distance. Nevertheless, results obtained by PRDC are not as accurate as the ones obtained by applying the more reliable NCD: while the latter is a relation between compression factors, the former is basically a compression factor in itself, and fails at normalizing according to the complexity of each object the similarity indices obtained [17].

## 3. FAST COMPRESSION DISTANCE

The Fast Compression Distance (FCD) is derived by combining the speed of PRDC without skipping the joint compression step which yields better performance with NCD. The idea is the following: a dictionary *D*(*x*) is extracted with the *LZW* algorithm [18] from each object encoded into a string *x*, and sorted in ascending order: the sorting is performed to enable the binary search of each pattern within *D*(*x*) in time $O(logN)$, where *N* is the number of entries in *D*(*x*). The dictionary is then stored for future use: this procedure may be carried out offline and has to be performed only once for each data instance. Whenever a string *x* is checked against a database containing *n* dictionaries, *D*(*x*) is extracted from *x* and matched against each of the *n* dictionaries. We define the FCD between *x* and an object *y* represented by *D*(*y*) as:

$$FCD(x, y) = \frac{|D(x)| - \bigcap(D(x), D(y))}{|D(x)|} \quad (2)$$

where $|D(x)|$ and $|D(y)|$ are the sizes of the relative dictionaries, represented by the number of entries they contain, and $\bigcap(D(x), D(y))$ is the number of patterns which are found in both dictionaries. A graphical representation of the mentioned sets is reported in Fig. 1. The $FCD(x,y)$ ranges for every $x$ and $y$ from 0 to 1, representing minimum and maximum distance, respectively. If $x = y$, then $FCD(x,y)=0$. Every matched pattern counts as 1 regardless of its length: the difference in size between the matched dictionary entries is balanced by *LZW*'s prefix-closure property which applies to the patterns contained in the dictionary. So, a long pattern $p$ common to $D(x)$ and $D(y)$ will be counted $|p|-1$ times, where $|p|$ is the size of $p$. The intersection between dictionaries represents the joint compression step performed in *NCD*, as the patterns in both the objects are taken into account.

*3.1 Speed comparison with NCD*

We can compare the numbers of operations needed by NCD and FCD to perform the most time-consuming step, and the only one which has to be repeated for the computation of every distance. This is the compression of the joint file $C(x, y)$ for NCD, and the computation of the intersection between the two dictionaries $D(x)$ and $D(y)$ for FCD. For sake of comparison, we consider LZ-based compression for NCD. The numbers of operations needed for two strings $x$ and $y$ are proportional to:

$$\begin{aligned}FCD(x, y) &\rightarrow \bigcap(D(x), D(y)) = m_x \log m_y \\ NCD(x, y) &\rightarrow C(x, y) = (n_x + n_y) \log(m_x + m_y)\end{aligned} \quad (3)$$

where $n_x$ is the number of elements in $x$ and $m_x$ the number of patterns extracted from $x$. In the worst case FCD is 4 times faster than NCD, if $x$ and $y$ have comparable complexity and are totally random. As regularity within an object $x$ increases, $m_x$ decreases with respect to $n_x$, as fewer longer patterns are extracted, and the number of operations needed by FCD is reduced. Other ideas can be used to further speed-up the computation. If in the search within $D(y)$ a pattern $p_x$ in $D(x)$ gives a mismatch, all patterns with $p_x$ as a prefix may be directly skipped: LZW's prefix-closure property ensures that they will not be found in $D(y)$. Furthermore, short patterns composed of two values may be regarded as noise and ignored if the dictionaries are large enough. The dictionaries extraction step may be carried out offline for FCD, therefore each dictionary needs to be computed only once for each object and can be then reused. In the average case, the experiments contained in this section will show that the complexity decreases by one order of magnitude even when we ignore every restriction on buffer size and lookup tables imposed by real compressors; this is done to expense of the generality of NCD, which is directly applicable to general data without a previous step of encoding the objects into strings.

## *3.2 CBIR System*

Typical CBIR systems operate with parameters representing the direct data content (typically color histograms, layouts, shapes, and/or invariant color features); in a classical query-by-example system, the user is able to present to the system a query image, and retrieve images which are similar, according to given criteria [20-22]. In this paragraph we define a system with these characteristics based on FCD.

Before extracting the dictionaries and computing the distance between images, it is needed to assign a single value to each pixel and convert the 2D image in a 1D string. As a first step, being the RGB channels correlated, the Hue Saturation Value (HSV) is chosen as color space, in order to have independent information in each channel.

A uniform quantization of the color space is then performed to avoid a full representation of the data, as using a limited alphabet facilitates the compressor in individuating shared patterns. In the HSV color space a finer quantization of *hue* is recommended with respect to *saturation* and *intensity*, since the human visual perception is more sensitive to changes in the former [23]. We use then 16 levels of quantization for the *hue* component, and 4 for both *saturation* and *value*. Therefore, the HSV color space is quantized in 8 bits, which allow a representation with $16 \times 4 \times 4 = 256$ values.

The images are then converted into strings: this introduces a loss of information, as traversing the image in raster order destroys the vertical texture, while the horizontal one is implicitly kept. Therefore we choose to add an extra bit to each pixel value, capturing the basic vertical interactions of the pixel. We assign 0 to smooth and 1 to rough transitions respectively between a pixel and both its adjacent vertical neighbours (see Fig. 2). For a pixel $p$ at row $i$ and column $j$, the value of the bit related to the vertical information $v_{i,j}$ is given by the following equation:

$$v(p_{i,j}) = \begin{cases} 1, & if \quad (d(p_{i,j}, p_{i+1,j}) > t) \, || \, (d(p_{i,j}, p_{i-1,j}) > t) \\ 0, & otherwise \end{cases}, \quad (4)$$

where

$$d(p1, p2) = \sqrt{\|h_{p1} - h_{p2}\|^2 + \|s_{p1} - s_{p2}\|^2 + \|i_{p1} - i_{p2}\|^2}, \quad (5)$$

$t$ is a threshold comprised between 0 and 1, and $h_p, s_p$ and $i_p$ are respectively the hue, saturation and intensity values of $p$ scaled from 0 to 1. In other words, it is simply checked whether the Euclidean distance in the HSV space between the norms of a pixel and any adjacent vertical neighbor is above a given threshold $t$. The decision to represent textural information with only one bit is adopted to keep a reduced alphabet size, to avoid hindering the compressor in finding similar patterns in the data. While this additional information improves the similarity indices obtained, multiple trials revealed that any more detailed textural information yields a worse performance, caused by an overfitting of the image content description.

According to [39], the local gradients in natural images are to some degree invariant from the image scale, therefore a single threshold can be chosen for all the images datasets at hand. This threshold $t$ should be chosen in order to convey the maximum information in the vertical texture (4), i.e. to yield $\max_t \{H(v) = -\sum_i p(i) \log p(i)\}$, where $H(v)$ is the entropy of $v$, measured over the values $i = \{0,1\}$ [40]. The entropy $H(v)$ is measured for different threshold settings on a set of test images coming from different datasets and classes, and the $\max_t \{H(v)\}$ is given for $t = 0.4$ An inspection confirmed that setting $t = 0.4$ for all the datasets taken into consideration splits the data in two sets of comparable cardinality.

Each image $x$ goes through the above steps of data preparation, is converted into a string by recurring the image in raster order, and is finally compressed, yielding a dictionary $D(x)$. If it is desired to retrieve images in the database which are similar to a query image $q$, one may apply a simple threshold to the *FCD* between $q$ and any object in the dataset and retrieve all the images within the chosen degree of similarity. A sketch of the workflow is depicted in Fig. 3.

## 4. EXPERIMENTS

In the experiments presented in this section we used the following datasets:

1. A subset of the COREL dataset [25], for a total of 1500 images equally divided in 15 classes, of which a sample is reported in Fig. 4.
2. The *Nister-Stewenius* (N-S) dataset [26], containing 2,550 objects, each of which is imaged from four different viewpoints, for a total of 10,200 images.
3. The *Lola* dataset [27], composed of 164 video frames extracted at 19 different locations in the movie "Run, Lola, run".
4. The *Fawns and Meadows* dataset [28], comprised of 144 meadows images, some of which contain a fawn hiding in the grass.

As scoring measures we used standard Precision-Recall, where Precision is the number of relevant documents retrieved by a query divided by the total number of documents retrieved, and Recall is the number of relevant documents retrieved divided by the total number of relevant documents [30]. In some experiments we also used general classification accuracy, and *ad hoc* scores for the N-S and *Lola* datasets; the variety of figures of merit used is for sake of comparison to distinct methods adopted in previous works which have performed experiments on these datasets. All experiments have been run on a machine with a double 2 GHz processor and 2GB of RAM.

*4.1 The COREL dataset*

In recent years content-based image retrieval systems relying on Vector Quantization (VQ) [31] have been defined. These are of particular interest for the scope of this work, as VQ is naturally related to data compression. Among these, the Minimum Distortion Image Retrieval (MDIR) by Jeong and Gray outperforms previous techniques based on histogram matching, by fitting to the training data Gaussian Mixture Models, later used to encode the query features and to compute the overall distortion [24] [32]. Daptardar and Storer introduce then a similar approach using VQ codebooks and mean squared error (MSE) distortion, decoupling to some degree spectral and spatial information by training separate codebooks in different regions of the images, outperforming in turn MDIR: we refer to their methodology as Jointly Trained Codebooks (JTC) [33].

Both MDIR and JTC have been tested on the COREL dataset with Precision vs. Recall curves, and in the following experiments the same set of 210 query images used by their authors has been considered, in order to have a fair comparison. All images of original size 256x256 have been resampled to different resolutions, from 128x128 to 32x32. This has been done considering the experiments contained in [34], where it is empirically shown that for a given 256x256 image is usually enough for a human to analyze its 32x32 subsampled version to understand the image semantic content and to recognize almost every object within the scene. We then compared the results for the same images with sizes of 128x128, 64x64 and 32x32 pixels. A slightly better performance has been obtained with the 64x64 images, with Fig. 5 showing the slight differences when adopting a different image size. Fig. 6 reports a comparison of FCD with MDIR and JTC: for values of recall higher than 0.2, FCD outperforms the previous techniques. In addition to a simple UQ a more refined VQ has been tested, with the training vectors being computed on the basis of 24 training images, but this representation did not improve the results. In addition, adopting a non uniform VQ would require a new computation of the vector quantizer whenever new semantic classes are added to the dataset.

A simple classification experiment has been then performed on the same dataset, with each image $q$ being used as query against all the others. For each query, $q$ has been assigned to the class minimizing the average distance: results obtained, reported in Table 1, show an accuracy of 71.3%. It has to be remarked that intraclass variability in the COREL dataset may be high: for example most of the 10 images not recognized for the African class may be in fact considered as outliers as just landscapes with no human presence are contained within (see Fig. 7); this shows the existence of limits imposed by subjective choices of the training datasets. The total running time for extracting the dictionaries and compute the distance matrix for the 1500 64x64 images is around 15 minutes, while it takes more than 150 with *NCD* (estimated on a 200 images dataset subset through the tool Complearn [35]).

*4.2 "Lola" dataset*

A sample of the *Lola* dataset is reported in Fig. 8. The retrieval performance is measured using the Average Normalized Rank *ANR* of relevant images [36] given by

$$ANR = \frac{1}{NN_r}\sum_{i=1}^{N_r} R_i - \frac{N_r(N_r+1)}{2} \tag{6}$$

where $N_r$ is the number of relevant images for a given query image, *N* is the size of the image set, and $R_i$ is the rank of the *i*th relevant image. The *ANR* ranges from 0 to 1, with the former meaning that all $N_r$ images are returned first, and with 0.5 corresponding to random retrieval.

In this case the results, reported in Fig. 9 and Table 2, are much worse than the best obtained by Sivic and Zissermann using Scale-Invariant Feature Transform (SIFT) methods [27]. Nevertheless they are acceptable, if we consider that no feature has been extracted from the scenes and no parameter had to be set or adjusted, and consistent with the Precision vs. Recall curve in the information retrieval experiment in Fig. 14.

*4.3 An Application to a Large Dataset: Stewenius-Nister*

A sample of the N-S dataset is depicted in Fig. 10. The measure of performance defined by the authors is counting how many of the 4 relevant images are ranked in the top-4 retrieved objects when an image *q* is used as query against the full or partial dataset.

Even though there would be faster query methods, to keep unaltered the workflow used so far we extracted all the dictionaries from the images and computed a full 10200x10200 distance matrix using the FCD as distance measure; afterwards, we checked the 4 closest objects for each image. To the best of our knowledge, this is the first time that a full distance matrix using compression-based similarity measure is computed on a dataset of this size. While this has been possible for the FCD in approximately 20 hours, the NCD would have required about 10 times more, so we built with the latter in 3 hours a 1000x1000 distance matrix related to a partial dataset, in order to compare performances. Results reported in Fig. 11 show that the FCD yields results as good as the NCD on the partial dataset, but not as good as the best obtained by Stewenius and Nister using SIFT methods; on the other hand, different combination of parameters and training sets yield very different results in the experiments of Stewenius and Nister, of which only some are better than the performance given by the FCD. For example, if the authors compute the score at one level only, namely on the leaves level of the hierarchical vocabulary tree adopted, results are slightly worse than the ones obtained by the FCD. Keeping in mind that more than 4 millions parameters are extracted in [37], while this step is skipped by the FCD, it can be notice how the latter avoids the drawbacks of working with algorithms in which the definition and setting of parameters plays a central role. Furthermore, the FCD does not adopt any *ad hoc* procedure for the dataset, but it is applied with no variations with respect to the other experiments contained in this section.

*4.4 Detection of wild animals*

Detection may be regarded as a subset of the classification task; about detection in images, in general the interest lies in knowing which images contain a certain object, or where the object is to be found within the images.

We tested the FCD in a wild animal detection experiment: the *Fawns and Meadows* dataset by Israel et al. [28] is comprised of 144 infrared images of meadows, 41 of which contain a fawn hiding in the grass; the image size is 160 x 120. The detection of fawns can be a tricky task, as patches of bare earth may have shape and temperature similar to the animals (Fig. 12). After the extraction of the dictionaries, as in the workflow in Fig. 3, the images have been classified on the base of their average distance from a class (fawn/meadows), with an accuracy of 97.9%, with 3 missed detections and 0 false positives, clearly outperforming NCD running with default parameters in both running time and accuracy (see Fig. 13 and Table 3). The processing of the full dataset (dictionaries extraction, distance matrix computation and decision process) took less than one minute. A Precision vs. Recall curve is reported in Fig. 14.

## 5. CONCLUSIONS

This paper introduced a similarity measure based on compression with dictionaries directly extracted from the data, the Fast Compression Distance (FCD). The FCD extracts offline a dictionary for each object which has previously been encoded into a string. In the string encoding step, some textural information is embedded within each pixel value to preserve as much information as possible. Subsequently, similarities between two objects are computed through an effective binary search on the intersection set between the relative dictionaries. The FCD has a reduced computational complexity with respect to the most popular compression-based similarity measure, the Normalized Compression Distance (NCD), as the latter processes iteratively the full data in order to discover similarities between the objects. At the same time, the data-driven approach typical of NCD is maintained, thus keeping a workflow with a parameter-free flavour.

Compression-based techniques can be then tested for the first time on medium-sized datasets composed of up to 10,000 objects, while in the past such methods have been usually applied to sets comprised of up to 100 objects, thus estimating their behaviour in a more statistically meaningful way. The FCD could represent a first step towards tackling the practical problems which arise when compression-based techniques have to be applied in data mining applications, as the query time for a dataset comprising more than 10,000 images was of 8 seconds on a standard machine, and could be further improved by optimizing the dictionary representation, for example by making use of built-in functions in Database Management Systems.

The experiments also show that FCD do not lose in accuracy with respect to NCD and often yields better performances, and we justify this with two remarks: firstly, the FCD should be more robust since it focuses exclusively on meaningful patterns, which capture most of the information contained in the objects; secondly, the use of a full dictionary allows discarding any limitation that real compressors have concerning the size of buffers and lookup tables employed, being the size of the dictionaries bounded only by the number of relevant patterns contained in the objects.

On the other hand, performances by fine-tuned systems based on invariant color features analysis often outperform FCD: this shows that compression-based techniques are not magic wands which yield in most cases the best results with minimum human intervention and therefore effort, as experiments on restricted datasets may have hinted in the past. Nevertheless, the overall highly satisfactory performance of these techniques, along with their universality, the simplicity in their implementation, and the fact that they require basically neither setting of parameters nor any supervision from an expert, justify the use of these notions in practical applications.

A distance similar to the FCD, the Normalized Dictionary Distance (NDD) has been independently proposed in [19], and proved to be a metric: the present paper can be seen as a companion of this work, as it derives from different considerations and draws different conclusions. Firstly, this paper considers the computational complexity aspects of the introduced distance, as it derives by combining the accuracy of NCD with the reduced complexity of PRDC. Secondly, it preserves basic textural information in the step of image encoding into strings. Finally, the proposed distance is validated on a larger number of datasets, and compared to other compression-based methods and state of the art techniques based on invariant color features.

With respect to traditional image analysis methods, the FCD and in general compression-based measures stand out as a different kind of image analysis, since patterns belonging to objects or to the background within the data are treated equally, being no part of the image privileged in the analysis. The FCD would then be more apt at capturing general image mood similarity rather than be focused on specific objects: depending on the applications, this could be an advantage or a drawback. On the one hand, the FCD is less suitable for the detection of objects within a scene; on the other hand, applications for which the context of an image is relevant would benefit from employing this technique: these include image types where the context analysis is crucial, such as satellite images, and applications to near copy detection.

A last remark has to be made on the lossless compression, adopted in order not to lose the universality of compression-based similarity measures. Lossy compression is the dominant form in multimedia and image representation, and is naturally connected to classification and retrieval. Therefore, the performance of lossy compression should be extensively tested: the dictionaries or codebooks extracted from the images could be compared through distortion measures to find the minimum distortion match to an observed signal. This could help in better capturing the relevant information within a given image, and would enable more complex comparisons between objects, also taking into account the frequency domain.

## ACKNOWLEDGMENT

The authors would like to thank prof. T. Watanabe for valuable suggestions and comments, and A. H. Daptardar, J. Sivic, and M. Israel for providing the datasets. Parts of Sections 3 and 4 were presented in [38].

# FIGURES

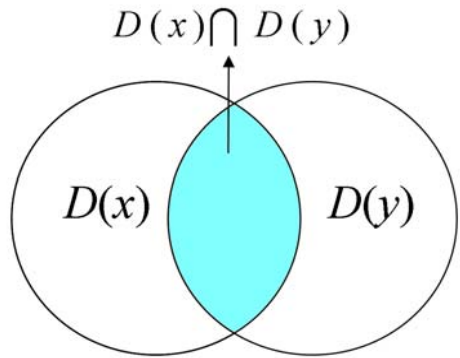

Fig. 1. Graphical representation of the intersection between two dictionaries $D(x)$ and $D(y)$, respectively extracted from two objects x and y through compression with the LZW algorithm.

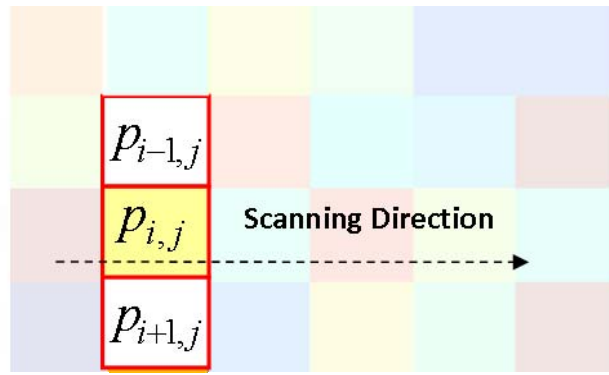

Fig. 2. Pixels considered to embed the basic vertical interactions information for pixel $p_{i,j}$ at a row $i$ and column $j$. A value of 0 and 1 is assigned to $p_{i,j}$ if the vertical texture is smooth or rough, respectively. Horizontal texture is not considered, as it is implicit in the compression step as the image is converted into string by traversing it in raster order.

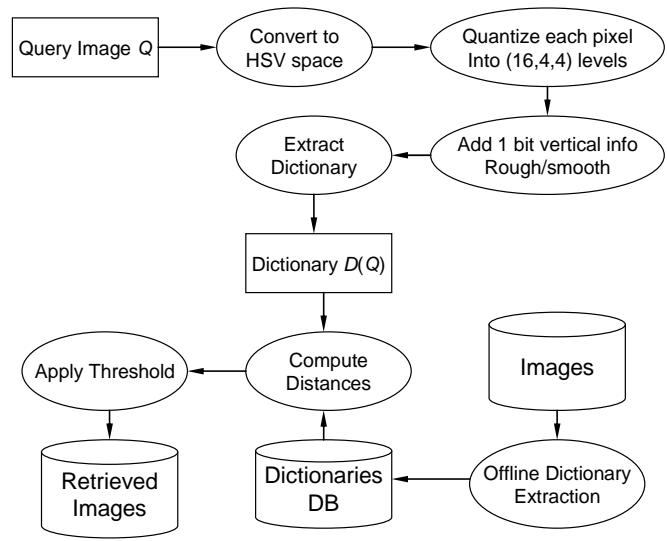

Fig. 3. Workflow for the dictionary-based retrieval system. After preprocessing, a query image *Q* generates a dictionary which is then compared to other dictionaries previously extracted from all the data instances.

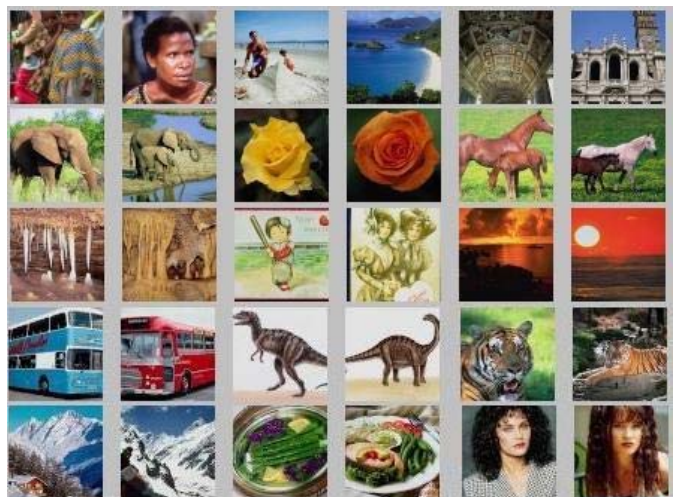

Fig. 4. Dataset sample of each of the 15 classes in raster order (2 images per class): Africans, Beach, Architecture, Elephants, Flowers, Horses, Caves, Postcards, Sunsets, Buses, Dinosaurs, Tigers, Mountains, Food, and Women.

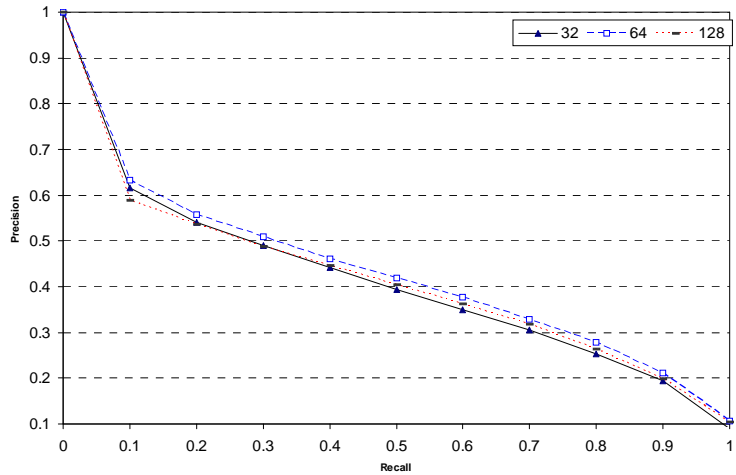

Fig. 5. Precision vs. Recall for different sizes of the images. The best performance is given for an image size of 64x64 pixels.

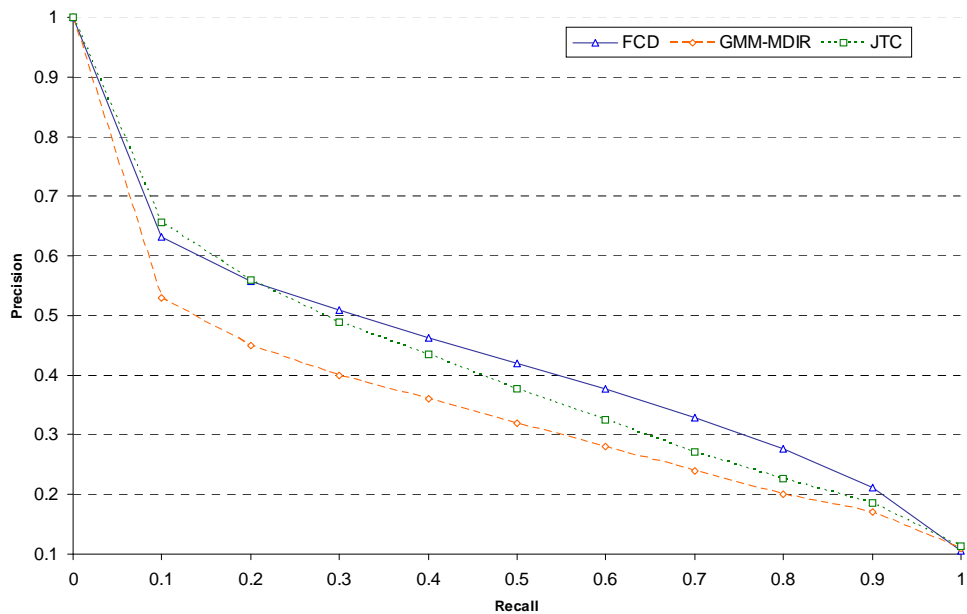

Fig. 6. Precision vs. Recall curves, comparing MDIR and JTC with the proposed method FCD. The FCD is applied after to the pixel values represented in the HSV color space, with an extra bit added to capture the essential vertical information, and after a scalar quantization of the HSV space.

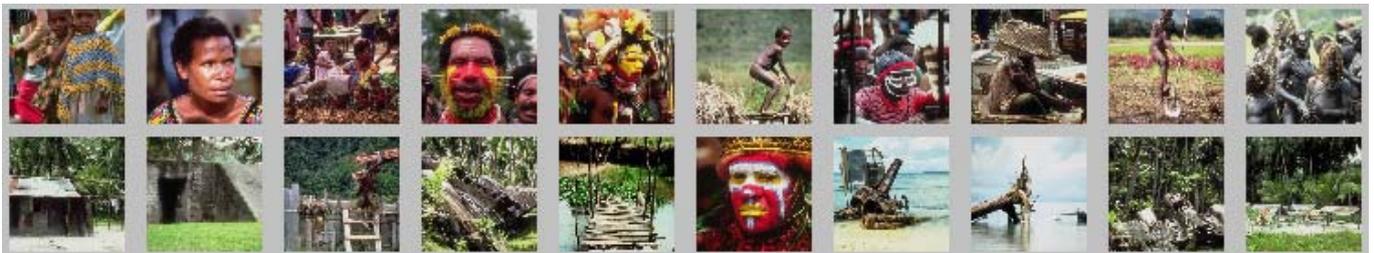

Fig. 7. Typical images for the class "Africans" (top row) and all misclassified images (bottom row), ref. Table 1. The false alarms depend on a subjective choice of the images, and the confusion with the class "tigers" is justified by the landscapes dominating the scenes with no human presence. An exception is represented by the 6th image in the bottom row (incorrectly assigned to the class "food").

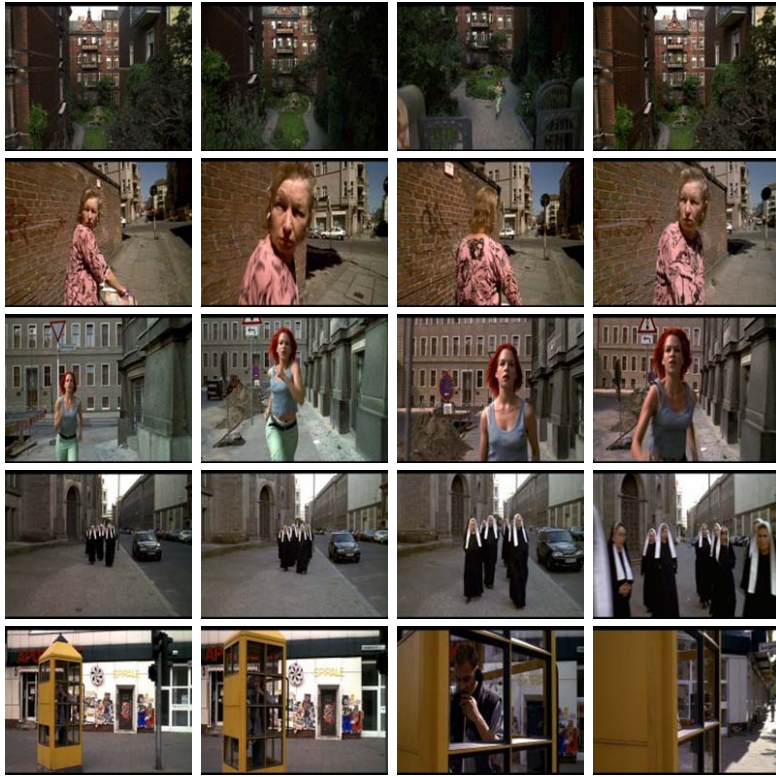

Fig. 8. Sample of Lola dataset. Each of the 5 rows contains images from the same class.

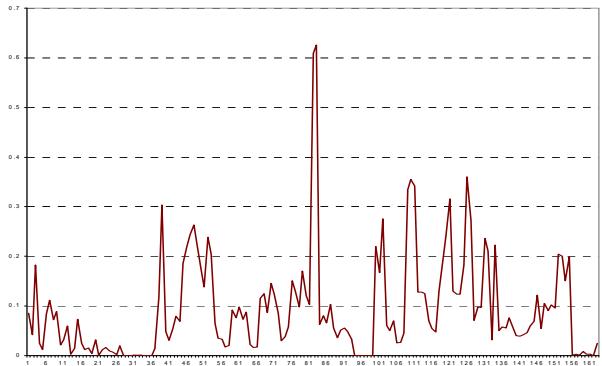

Fig. 9. Lola dataset *ANR* score for each object in the dataset. The values of 0 and 0.5 for each object correspond respectively to exact and random retrieval of all the images in the same class in the dataset.

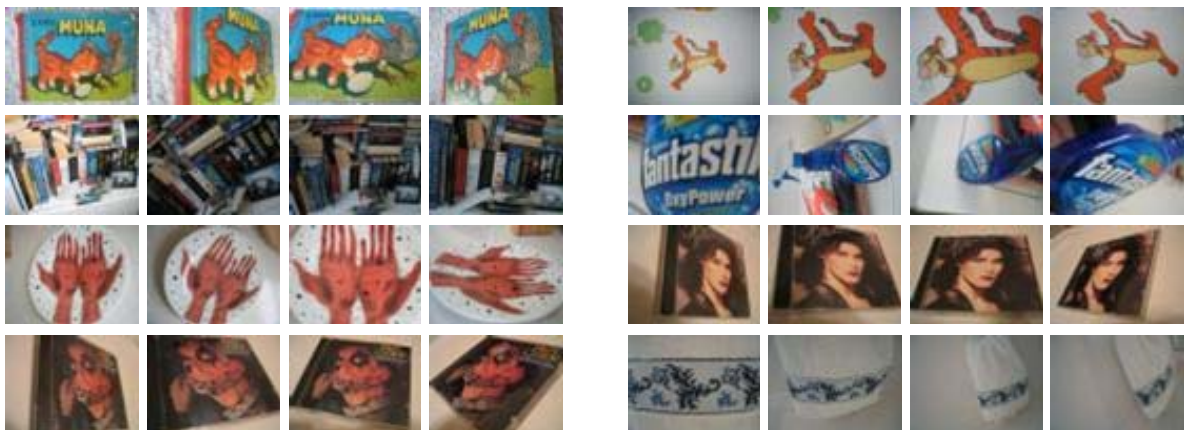

Fig. 10. Sample of Nister-Stewenius dataset.

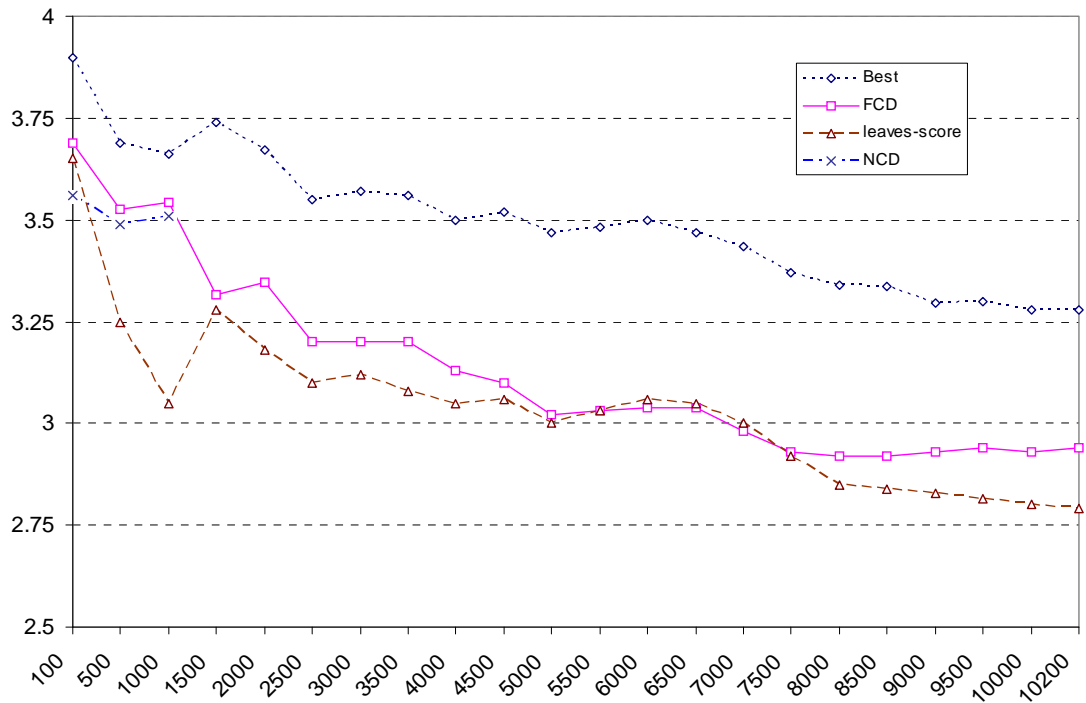

Fig. 11. Stewenius-Nister dataset score, upper-bounded by 4. Comparison with SIFT-based methods with different parameters settings, and comparison with NCD for a partial dataset. The *x* axis shows the size of the dataset subset considered.

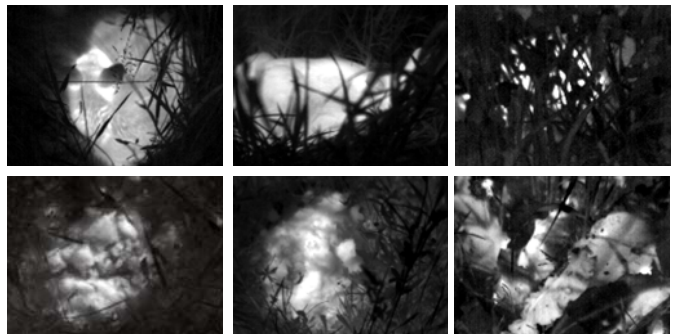

Fig.12. Sample from the *Fawns and Meadows* dataset. Top row: fawns. Bottom row: meadows.

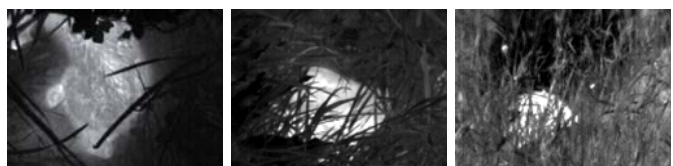

Fig. 13. The 3 fawns not detected by FCD (ref. Table 3).

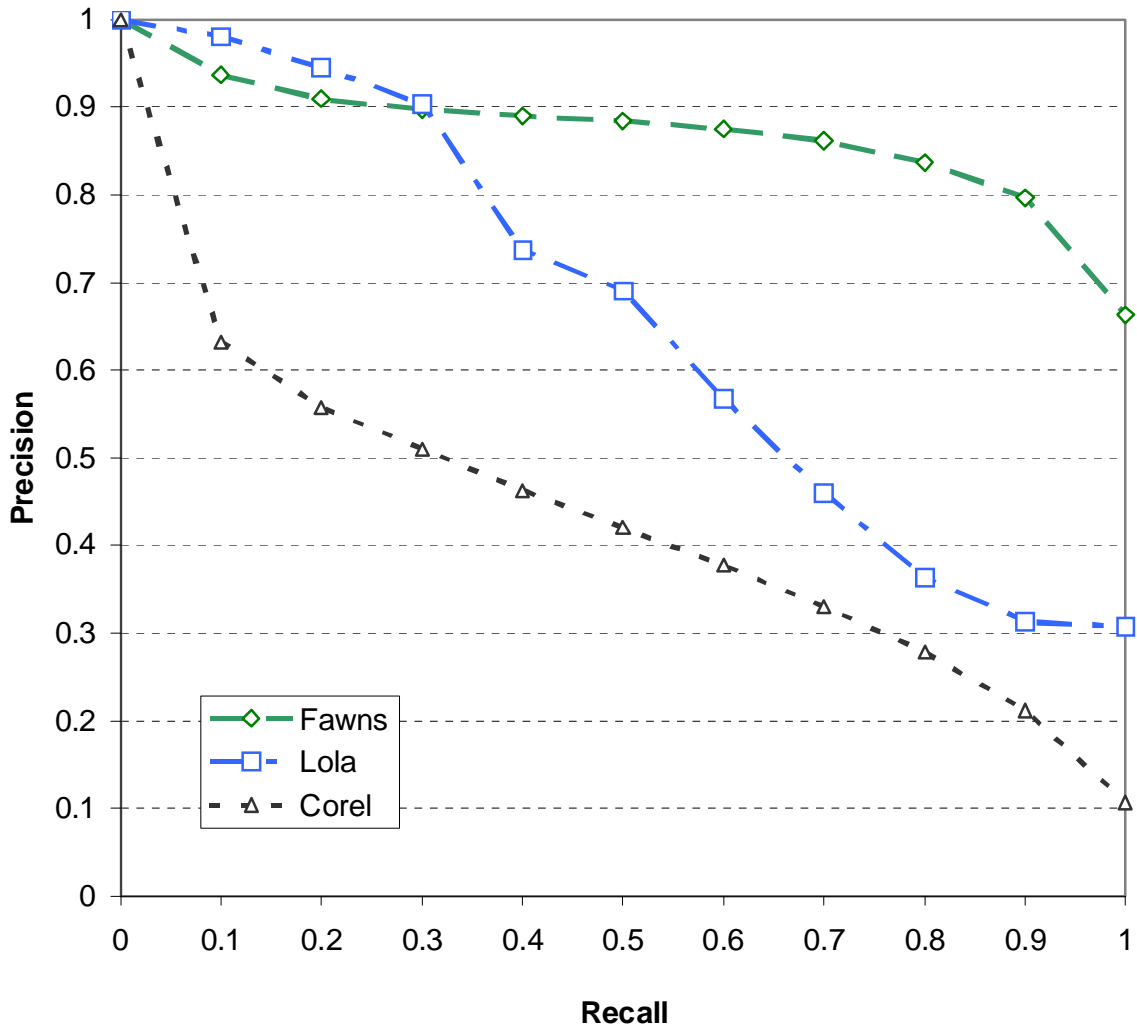

Fig. 14. Precision-Recall curves for most of the datasets analyzed in this work. The comparison has the only purpose of estimating which dataset presents more difficulties in being categorized. A dataset characterized by greater content diversity yields a lower curve in the graph.

# TABLES

TABLE 1
CONFUSION MATRIX FOR CLASSIFICATION ACCORDING TO THE MINIMUM AVERAGE DISTANCE FROM A CLASS

|  | Afr. | Beach | Archit. | Bus. | Dinos. | Eleph. | Flow. | Hors. | Mount. | Food | Caves | Post. | Suns. | Tig. | Wom. |
|---|---|---|---|---|---|---|---|---|---|---|---|---|---|---|---|
| Africans | **90** | 0 | 0 | 0 | 1 | 0 | 0 | 0 | 0 | 1 | 0 | 0 | 0 | 8 | 0 |
| Beach | 12 | **43** | 8 | 14 | 0 | 1 | 0 | 0 | 1 | 3 | 0 | 0 | 0 | 18 | 0 |
| Architecture | 7 | 0 | **72** | 3 | 0 | 0 | 0 | 0 | 0 | 1 | 0 | 0 | 1 | 16 | 0 |
| Buses | 6 | 0 | 0 | **93** | 0 | 0 | 0 | 0 | 0 | 1 | 0 | 0 | 0 | 0 | 0 |
| Dinosaurs | 0 | 0 | 0 | 0 | **100** | 0 | 0 | 0 | 0 | 0 | 0 | 0 | 0 | 0 | 0 |
| Elephants | 16 | 0 | 2 | 2 | 0 | **46** | 0 | 4 | 0 | 3 | 0 | 1 | 0 | 26 | 0 |
| Flowers | 6 | 0 | 3 | 1 | 0 | 0 | **83** | 1 | 0 | 3 | 0 | 0 | 0 | 3 | 0 |
| Horses | 0 | 0 | 0 | 0 | 0 | 0 | 0 | **97** | 0 | 0 | 0 | 0 | 0 | 3 | 0 |
| Mountains | 7 | 1 | 11 | 23 | 0 | 2 | 0 | 0 | **39** | 0 | 0 | 0 | 0 | 17 | 0 |
| Food | 6 | 0 | 0 | 1 | 0 | 0 | 0 | 0 | 0 | **92** | 0 | 0 | 0 | 1 | 0 |
| Caves | 17 | 0 | 9 | 1 | 0 | 1 | 0 | 0 | 0 | 5 | **60** | 0 | 0 | 7 | 0 |
| Postcards | 0 | 0 | 0 | 0 | 1 | 0 | 0 | 0 | 0 | 1 | 0 | **98** | 0 | 0 | 0 |
| Sunsets | 18 | 0 | 1 | 6 | 0 | 0 | 2 | 0 | 0 | 16 | 3 | 1 | **39** | 14 | 0 |
| Tigers | 1 | 0 | 0 | 1 | 0 | 0 | 0 | 5 | 0 | 0 | 0 | 0 | 0 | **93** | 0 |
| Women | 35 | 0 | 0 | 6 | 2 | 0 | 0 | 0 | 0 | 20 | 4 | 0 | 0 | 5 | **28** |
| **Avg Accuracy** | | | | | | | 71.3% | | | | | | | | |

TABLE 2
ANR SCORES FOR THE LOLA DATASET

| FCD | 0.093 |
|---|---|
| SIFT | 0.013 |

TABLE 3
CONFUSION MATRIX FOR THE FAWNS DATASET

|  |  | Fawn | Meadow | Accuracy | Time |
|---|---|---|---|---|---|
| **FCD** | Fawn | 38 | 3 | 97.9% | 58 sec |
|  | Meadow | 0 | 103 | | |
| **NCD** | Fawn | 29 | 12 | 77.8% | 14 min |
|  | Meadow | 20 | 83 | | |